\documentclass{llncs}
\usepackage{indentfirst,amssymb,amsmath,latexsym,graphicx}

\title{Machine Learning with Probabilistic Law Discovery: A Concise Introduction}

\author{Alexander~Demin and Denis~Ponomaryov}
\institute{Ershov Institute of Informatics Systems, Novosibirsk, Russia \vspace{0.2cm}\newline \email{alexandredemin@yandex.ru, ponom@iis.nsk.su}}

\begin{document}
\maketitle

\begin{abstract}
Probabilistic Law Discovery (PLD) is a logic based Machine Learning method, which implements a variant of probabilistic rule learning. In several aspects, PLD is close to Decision Tree/Random Forest methods, but it differs significantly in how relevant rules are defined. The learning procedure of PLD solves the optimization problem related to the search for rules (called \emph{probabilistic laws}), which have a minimal length and relatively high probability. At inference, ensembles of these rules are used for prediction. Probabilistic laws are human-readable and PLD based models are transparent and inherently interpretable. Applications of PLD include classification/clusterization/regression tasks, as well as time series analysis/anomaly detection and adaptive (robotic) control. In this paper, we outline the main principles of PLD, highlight its benefits and limitations and provide some application guidelines. 
\end{abstract}


\section{Introduction}\label{Sect:Intro}
Despite the popularity of neural network based and boosting models, there is still a big interest to logic based methods of Machine Learning, which support explicit representation of learned hypotheses. The inherent interpretability of these methods is what makes them particularly useful in critical domains such as, e.g., information security, medicine, automated control, etc.

Of particular interest in the field of logic based Machine Learning is Probabilistic Law Discovery (PLD) \cite{b2}, which is a variant of probabilistic rule learning. It allows for balancing between the completeness of the set of the learned hypotheses and computational expenses, and in the limit it guarantees learning the complete set of hypotheses true on data.

While having some similarities with Decision Tree/Random Forest methods, PLD based models uniquely combine ensembling features with the property of being inherently interpretable. The explicitness of hypotheses learned by PLD allows for building glass-box classification, clusterization, regression, or adaptive control models, which also support straightforward integration of domain knowledge. The transparency of PLD based models makes them accessible for post-hoc meta-analysis to support transfer learning, conceptual abstraction, symmetry detection, etc. Similarly to some other logic based ML methods, the disadvantages of PLD are due to the complexity of rule learning, which is related to NP-hard problems and thus, direct implementations of PLD face efficiency problems when applied to datasets with big numbers of features. 

Currently several implementations of PLD are known, which combat the dimensionality problem with the help of heuristics. They have been benchmarked on different ML tasks against other well-known models, e.g., decision tress, neural networks, associative rules, in domains such as medicine \cite{b16}, finance \cite{b13,b17,b18}, bioinformatics \cite{b14}, adaptive control \cite{b12,b8,b11,b9,b7}.

The aim of this paper is to provide a concise and accessible introduction into Probabilistic Law Discovery, which covers the base learning algorithms, optimization techniques, and application guidelines. The exposition is based on the latest implementation of PLD, which provides a reasonable balance between the completeness of the learned hypotheses and computational complexity. 

\section{Principles of Probabilistic Law Discovery}
Probabilistic Law Discovery is based on learning probabilistic rules on data as expressions in a human-readable formal language. Conceptually close to PLD are the decision tree/random forest methods, but the main difference is in how the most informative rules are defined and how they are learned.

\subsection{Probabilistic Rules} 
A (probabilistic) rule is an expression of the form
\begin{equation}
P_1(x), \ldots , P_n(x) \rightarrow R(x)
\end{equation}
where $x$ is a variable and $R, P_1, \ldots , P_n$, $n \geqslant 0$, are predicates. $P_1(x), \ldots , P_n(x)$ is the \emph{premise} of the rule and $R(x)$ is the \emph{conclusion}. The rule \emph{size} is the number of predicates in the premise. It is a common requirement that all predicates must be ``simple enough'' to compute. In applications, this requirement is specified as the existence of a polynomial/linear/logarithmic algorithm (wrt the size of the dataset) to compute the predicates. For instance, if there is a predicate $HasChildren$ then there must be a procedure to compute in at most polynomial time (wrt the input data) for any object $o$ whether $HasChildren(o)$ holds on data.

In general, a predicate can be given as a formula of first order logic with one free variable (called the \emph{object variable}). Complex predicates (such as, e.g., $HasChildren(x) \vee HasBrother(x)$ or $\exists y \ Brother(x,y)$) can be written in terms of base predicates ($HasChildren$, $HasBrother$). In order to use PLD, one must first define a set of base predicates and procedures to compute them and then (if needed) one can introduce complex predicates as formulas over the base predicates.

The following are examples of rules:
\begin{align*}
CheapItem(x) & \rightarrow HighDemand(x)\\
HasChildren(x), \exists y \ Brother(x,y) & \rightarrow Daughter(x)
\end{align*}

For object-feature datasets, the base predicates can be taken as corresponding to the features. However, with complex predicates one can formulate more expressive rules. For example, the predicate $HasChildren(x) \vee HasBrother(x)$ gives the set of objects that have at least one of the features. The predicate $\exists y \ Brother(x,y)$ employs a relationship of objects. Depending on the choice of predicates we can see more or less information in the data.

For an object-feature dataset $\cal{D}=\langle \cal{O}, \cal{F}\rangle$, let $p$ be a probability measure on the set of objects $\cal{O}$. The \emph{probability} of a rule $P_1(x), \ldots , P_n(x) \rightarrow R(x)$ on $\cal{D}$ is defined as the value of $p(P_1, \ldots , P_n, R)$ divided by $p(P_1, \ldots , P_n)$, where $p(P_1, \ldots , P_n, R)$ is the probability measure of those objects $o\in\cal{O}$, for which all $P_1(o), \ldots , P_n(o), R(o)$ hold (the meaning of $p(P_1, \ldots , P_n)$ is defined similarly). In applications of PLD, the frequency probability measure is typically used, and thus, the rule probability reflects the number of objects for which both, the premise and conclusion hold, in relation to the number of objects, for which only the premise is true. 

\subsection{Probabilistic Law Learning}
Given a predicate language, the natural question is which rules best reflect the regularities hidden in the data. One can enumerate rules in a brute-force fashion and estimate their probabilities, but the number of all rules in the given language is large and for many rules the probability may be close to zero, which means they are not too informative. 

PLD is based on the assumption that important are those rules, which have a minimal size and relatively high probability. This corresponds to the classical trade-off between the size and informativeness of compressed data representation (the so called Minimum Description Length principle). Essentially, PLD is solving a certain optimization problem of minimizing rule length, while maximizing the probability. 

A rule $P_1(x), \ldots , P_n(x) \rightarrow R(x)$ is said to be a \emph{probabilistic law} on a given dataset if it has a non-zero probability $p$ and the following holds: the probability of any other rule with the same conclusion and a premise given by a proper subset of predicates $P_1, \ldots , P_n$, is strictly less than $p$. 

Thus, $P_1(x), \ldots , P_n(x) \rightarrow R(x)$ is the shortest rule (by inclusion of premises)  with conclusion $R$, which has probability $p$. Note that the definition leaves the possibility that there may exist a rule with a superset of predicates in the premise and with the same conclusion, for which the probability is greater than $p$.

It is known that the problem to find short rules having a given probability is computationally difficult. There may be exponentially many shortest rules, which have probability $1$ on data \cite{Kuznetsov}, and the problem to decide whether there exists one with at most $k$ predicates in the premise is NP-complete \cite{SpecificSentences}. This implies that it is hard to compute the set of all probabilistic laws on a given dataset. The learning procedure of PLD is implemented by a heuristic algorithm, which has the following properties:
\begin{itemize}
\item for a given dataset and a predicate language, it outputs a set of rules in this language, which are probabilistic laws on this dataset
\item in general the algorithm does not guarantee to find all the probabilistic laws on the dataset
\item it allows for balancing between the completeness of the obtained set of laws and computation time 
\item in the limit (having unbounded computational resources) the algorithm computes the complete set of probabilistic laws on the given data
\end{itemize}
 
The algorithm implements a directed enumeration. For a predicate $R$, it outputs a set of probabilistic laws with conclusion $R$. In practice, however, the computations can be organized in such a way as to obtain probabilistic laws for all conclusion predicates in one pass. The algorithm enumerates the rules with conclusion $R$ in a directed manner, starting from the rule with the empty premise ($\emptyset\rightarrow R$), by refinement (i.e., by adding predicates one-by-one to the premise). Clearly, refining a rule may  change its probability.

\begin{figure}[htbp]
\makebox{\centerline{\includegraphics[height=1.35in,keepaspectratio]{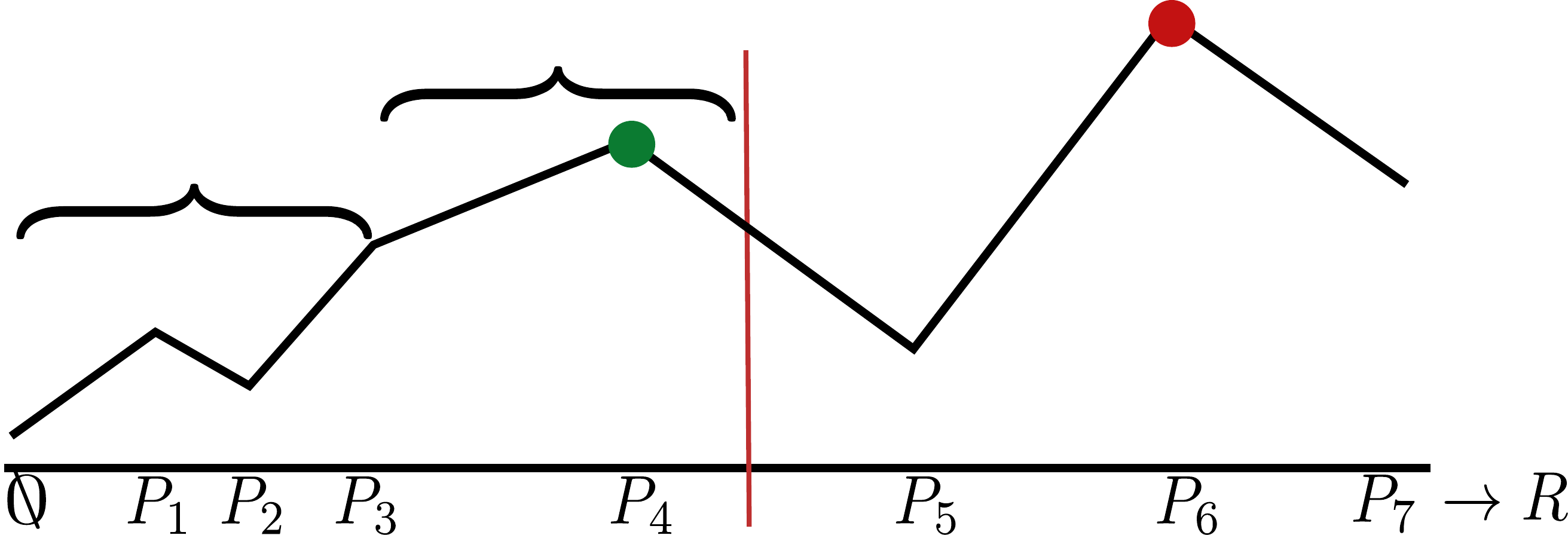}}
\put(-320,67){{\small base enumeration}} \put(-275,92){{\small additional enumeration}}}

\caption{Probability change for rules with conclusion $R$ and different premises}
\label{Fig:HillClimbing}
\end{figure}

The principle hyperparameter of the algorithm is the base rule enumeration depth (denoted as $d$). The algorithm enumerates all rules whose premise consists of at most $d$ predicates and selects those ones which are probabilistic laws. Since the base enumeration is complete, it is guaranteed that all probabilistic laws with at most $d$ predicates in the premise are computed.

After that, the algorithm selects from the obtained probabilistic laws those ones, which have exactly $d$ predicates in the premise, and it starts refining only these rules (by successively adding predicates one by one to the premises), while checking whether their probability increases (i.e., whether the resulting rule remains a law). This stage of the algorithm is called additional enumeration.

The premise of a rule is extended with new predicates as long as its probability increases. If, after an  addition of a predicate, the probability is not increased, then such a refinement of the rule is discarded. Figure \ref{Fig:HillClimbing} illustrates the situation when the algorithm computes the probabilistic law $P_1,\ldots,  P_4\rightarrow R$. When the predicate $P_5$ is added, the probability drops down, but as the predicate $P_6$ is added, it reaches a value that dominates the probabilities of all the rules with shorter premises (by set inclusion). That is, the rule $P_1,\ldots , P_6\rightarrow R$ is a longer probabilistic law. In this situation, the algorithm does not find the rule $P_1,\ldots , P_6 \rightarrow R$ at additional enumeration and thus, it does not guarantee to find global maxima (note the red point in Figure \ref{Fig:HillClimbing} vs the local maxima depicted by the green point). In this sense, the algorithm can be classified as a hill-climbing one.

The heuristic used in the algorithm is based on the assumption that probabilistic laws are typically arranged into chains, in which each subsequent law is obtained from the previous one by refinement with a single predicate, for example: $\{$ $P_1, P_2\rightarrow R\ \ $, $P_1, P_2, P_3\rightarrow R\ \ $, $P_1,P_2,P_3,P_4\rightarrow R$ $\}$. At the base enumeration step, the algorithm tries to capture the beginning fragments of these chains, and then, in the additional enumeration, it tries to find other laws from these chains.

\subsection{Requirement to Input Datasets}
The input data must be converted into a tabular object-feature representation. Categorical features must be converted to Boolean ones (e.g., by using one-hot encoding). Numeric features must be quantized and converted to Boolean ones, for example, by using an iterative splitting into ranges of values greater/smaller than the median.

\subsection{Application Scenarios}
PLD based models are used for the following tasks:
\begin{itemize}
\item probabilistic prediction of features (the classification task) \cite{b6,b16,b14}
\item identification of features/specific feature values that an object must have in order to be assigned to a particular class (abductive classification)
\item combining features into subsets closed wrt probabilistic laws and computing subsets of objects corresponding to these closed subsets (hierarchical object and feature clusterization) \cite{b3,b4,b5}
\item prediction of value intervals for numeric features (interval regression) 
\item anomaly detection and time series analysis \cite{b15,b17,b18}
\item building self-learning agent systems that interact with environments (reinforcement learning) \cite{RL}
\item control of modular systems with many degrees of freedom, in particular, adaptive robotic control \cite{b12,b8,b11,b9,b7}
\end{itemize}

We comment on solutions to these tasks in Section \ref{Sect:Applications}. 

\section{The PLD Algorithm}
\begin{figure}[htbp]
\makebox{\centerline{\includegraphics[height=1.1in,keepaspectratio]{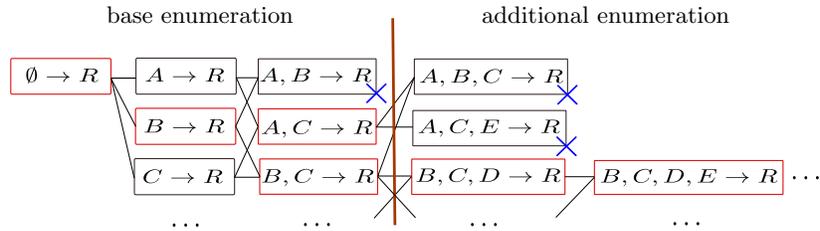}}
\put(-290,77){{\footnotesize base enumeration}} \put(-148,77){{\footnotesize additional enumeration}}}
\caption{An example rule derivation graph for base enumeration depth $d=2$}
\label{Fig:Graph}
\end{figure}

The implementation of the Probabilistic Law Discovery algorithm is based on the construction of a rule derivation graph. The nodes in this graph are the rules enumerated by the algorithm and there is an edge from a rule $r$ to $r'$ if $r$ is a subrule of $r'$, i.e. they have the same conclusion and the set of predicates from the premise of $r$ is a proper subset of predicates from the premise of $r'$. Figure \ref{Fig:Graph} illustrates an example derivation graph over some dataset and the predicate language $\{R, A,B,C,D,E,\ldots\}$. In this figure, only those edges are shown, which correspond to refinements computed by an (example) run the PLD algorithm.

Probabilistic laws in Figure \ref{Fig:Graph} are marked with the red color and those rules, which are not  refined, are marked with a cross. In particular, $A\rightarrow R$ is not a probabilistic law (i.e., it has a probability less or equal to the probability of $\varnothing\rightarrow R$). In this example illustration, all rules up to the depth $d=2$ are enumerated and it turns out that there are probabilistic laws $A, C\rightarrow R$ and $B, C\rightarrow R$ of depth $2$. At the additional enumeration phase, these rules are refined, thus giving the probabilistic law $B, C, D\rightarrow R$, which is refined further.

Using the graph structure to represent rules allows for optimizing the search for probabilistic laws in several ways. By using the graph, one can quickly find subrules of a given rule when checking the conditions for being a probabilistic law. Also the graph allows one to store the statistics on the calculated rules so as not to count them twice. Since the same rule can be a subrule of several other child rules, its statistics (the data to compute the probability and significance of the rule) can be reused.
 
Let $R$ be some target predicate (e.g., a feature to be predicted). To compute a set of probabilistic laws with conclusion $R$, the derivation graph is used by the PLD algorithm as follows. 

Initially, the root node of the graph is generated with the rule $\varnothing\rightarrow R$. Statistics for this rule, such as probability, confidence, etc., are computed. Auxiliary sets $REG_0$ and $Nodes_0$ are initialized to consist of this single rule.

At step $1\leqslant k\leqslant d$, where $d$ is the base enumeration depth ($d\geqslant 1$), the next graph level is built, which consists of the nodes obtained as the refinement of the rules from $Nodes_{k-1}$ with a single predicate. The resulting nodes is the set $Nodes_k$. For each node $N$ from $Nodes_k$:
\begin{itemize}
\item statistics for $N$ is calculated
\item $N$ is connected by an edge to each subrule of $N$ from the previous graph layer
\item it is verified whether $N$ is a probabilistic law; the set $REG_k$ is defined to consist of all probabilistic laws from $Nodes_k$, which meet a statistical criterion
\end{itemize}

At step $k > d$ (additional enumeration), the set of all single-predicate refinements of probabilistic laws from $REG_{k-1}$ is computed. The resulting nodes is the set ${Nodes}_k$. For each node $N$ from ${Nodes}_k$:
\begin{itemize}
\item it is checked whether $N$ meets a statistical criterion
\item subrules of $N$ size $k-1$ are searched in the previous graph layer by using $FindParents$ procedure as follows. If there is a subrule $r$, for which $p(r)\leqslant p(N)$, then $N$ is not a probabilistic law and the procedure stops. Otherwise, each found subrule $r$ is connected with node $N$ by an edge. If some subrule of node $N$ is not found in the previous layer, then a node $N_{sub}$ is created for this subrule in the graph and its statistics is calculated. If $p(N_{sub})\leqslant p(N)$ then the procedure stops. Finally, $FindParents$ is applied recursively to each subrule node of $N$.
\end{itemize}

The set $REG_k$ is defined to consist of the probabilistic laws $N$ which meet a statistical criterion.

\medskip

The PLD algorithm stops at a step $k \geqslant d$ (and it outputs the union of $REG_i$, for all $0 \leqslant i \leqslant k$) if either of the following conditions holds:
\begin{itemize}
\item the set $REG_k$ is empty (i.e., there are no probabilistic laws at level $k$ in the graph);
\item the number $k$ equals to the maximal rule size $MaxSize$ (a hyperparameter for setting the maximal number of predicates in premises of rules considered by the algorithm).  
\end{itemize}   

We note that the rule $\varnothing\rightarrow R$ in $REG_0$ provides some basic information about how likely the objects from the dataset are to have feature $R$ (under no further conditions). It is used, for example, in the classification task, in order to avoid prediction failure in cases where the attributes of an object being classified ``cover'' none of the premises of probabilistic laws with conclusion $R$.

Note also that probabilistic laws are filtered out wrt a statistical criterion. For example, it can be the case that there is a single object in a dataset with the feature $CheapProduct$ and this object has also the feature $HighDemand$. In this case the rule $CheapProduct(x)\rightarrow HighDemand(x)$ has probability $1$ on the data, but it applies only to the single object and therefore is not informative for generalization to new samples. Because of this the PLD algorithm estimates the statistical significance of probabilistic laws. In implementations of PLD a statistical significance test with a confidence interval $a$ is used, which is a hyperparameter of the algorithm.

To further restrict the number of considered rules, the algorithm implements several optimizations, which we discuss below. 

\subsection{Optimizations}
The PLD algorithm faces two principal computational problems. The algorithm implements a search in a rule space of size exponential wrt the number of predicates (features) from the given predicate language. Despite the additional enumeration heuristic, the number of rules considered by the algorithm may still be too large to complete computation within reasonable resources. The second problem is related to the estimation of rule statistics. For instance, computing rule confidence (the number of objects, for which the premise of a rule is true), when implemented naively, requires a full scan of the dataset.

To solve the first problem, the following additional criteria and hyperparameters are used to reduce the number of probabilistic laws considered at additional enumeration:

\begin{itemize}
\item \textbf{Probability threshold.} If the probability of a law is below a given threshold, then it is dropped (and is not further refined).
\item \textbf{Statistical significance threshold.} If the statistical significance of a probabilistic law is under a specified threshold, then it is dropped.
\item \textbf{Probability gain threshold.} Those refinements of a probabilistic law $r$ are dropped that give a probability gain (wrt the value of $p(r)$) less than a specified threshold.
\item \textbf{Probability gain thresholds for each level in the derivation graph.} Similar to the previous criterion, a probability gain threshold can be set separately for each level in the graph. This allows for restricting the number of laws specifically for each level, in case there is a blow-up in the number of laws at certain graph levels.
\end{itemize}

To solve the second computational problem, the derivation graph is used to speed up subrule search and statistics computation. In particular, the algorithm employs index caching for dataset objects: for a rule $r$, references to all those objects are stored, which $r$ is applicable to (i.e., on which the rule premise is true). Then to compute statistics for a refinement $r'$ of $r$ only these objects are used, one does not need to scan the complete dataset. This optimization significantly speeds up computations in practice. Clearly, the downside of caching is the need to store multiple indexes, which requires additional memory.

\section{Hyperparameters and Tuning}\label{Sect:Hyperparameters}
The following list summarizes the main hyperparameters of PLD:
\begin{itemize}
\item base rule enumeration depth $d$
\item maximum rule size (maximal enumeration depth) $MaxSize$
\item probability threshold for rules
\item confidence (statistical significance) threshold for rules
\item probability gain threshold (global threshold) for laws
\item probability gain thresholds for each level of the derivation graph and for each law size (level/size specific threshold)
\end{itemize}

One of the shortcomings of the current PLD implementation is related to the problem that it is hard to tell in advance how much time/memory will be required for searching probabilistic laws up to a given size $MaxSize$. It might happen that there are too many probabilistic laws at some enumeration level. When iterating over refinements of these laws at the next level one faces combinatorial explosion: it is impossible to complete the search within acceptable time or memory limit for storing the derivation graph and caching statistics.

One of the important aspects here is hyperparameter tuning, for which we recommend the following approach. One can make first a test run of PLD on a given training data with all thresholds set to zero. If an explosive growth in the number of rules is observed the thresholds are increased. Either the global probability gain threshold is increased or the level/size specific ones, if explosion occurs at a specific level of the derivation graph. The procedure is repeated until the algorithm is able to iterate to the specified size $MaxSize$ within acceptable time/resources.

Another way is to adjust hyperparameters wrt the quality of predictions based on the computed probabilistic laws. In situation when PLD cannot enumerate laws up to a given size $MaxSize$ we have a choice: either find shorter probabilistic laws by reducing $MaxSize$, or implement enumeration up to the required size by dropping some shorter laws rules by adjusting other thresholds. In this case, one can proceed in the standard way: select a small test subset of the training data and choose a option which provides the best prediction on this test subset.

\section{Applications}\label{Sect:Applications}
The rule learning approach implemented by PLD is employed in different ML tasks as follows. 
\subsection{Classification}
For each class label, a predicate (called \emph{predictor}) is introduced into the language and the PLD algorithm is run to learn probabilistic laws with these predicates in the conclusion. The resulting laws are used for classification of data objects in the following way. For an object, a subset of laws applicable to its features is selected. A law is applicable if the set the predicates from its premise is a subset of the predicates corresponding to the object features. Then the laws with maximal probability values are selected. If there is a single law of this kind then the class label for the object is defined by the predictor from the conclusion of the law. The probability for this label is defined as the probability value of the law. If there are several such laws then the object is not assigned a label (in this case classification fails). 

\subsection{Clusterization}
Object and predicate (feature) clusters are defined via probabilistic laws in the following way. For a subset of predicates $\cal{F}$ an \emph{agreement measure} of $\cal{F}$ is calculated as the difference between the sum of probabilities of laws $P_1(x), \ldots ,$ $P_n(x)$ $\rightarrow R(x)$, $n\geqslant 0$ s.t. $\{P_1, \ldots , P_n, R\}\subseteq \cal{F}$ and the sum of probabilities of those laws, whose premise is in $\cal{F}$, but the conclusion is not. The measure reflects the difference between the total probability of laws true on $\cal{F}$ and the total probability of laws false on $\cal{F}$. Predicate (feature) clusters are then defined via local maxima of the agreement measure: adding or removing any single predicate from a cluster yields a lower measure value. The set of laws true on a cluster $\cal{F}$ is called a  \emph{characteristic set} of $\cal{F}$. Then two objects are assigned to the same (object) cluster if their feature sets have similar agreement measures wrt a characteristic set of some feature cluster. The result of a PLD based clusterization obtained this way is a partially ordered (wrt set inclusion) hierarchy of feature and object clusters.    

\subsection{Regression}
In terms of PLD, regression is solved in several ways. The first one employs quantization: the rule language is extended with predicates, which correspond to certain ranges of feature values. The PLD algorithm is applied to learn probabilistic laws with these predictor predicates in the conclusion. Then regression is reduced to classification with the predictors being the class labels. Another approach employs averaging: if for an object $o$ we have $k\geqslant 1$ probabilistic laws with predictors $R_1, \ldots , R_k$ applicable to $o$ and the ranges, which correspond to predictors, support averaging, then the average value for these ranges is returned as the resulting value. The third approach employs rules with binary inequality predicates for interval regression. We provide here an illustrating example. In general, PLD can support rules with n-ary predicates and functional terms via grounding (we comment on this extension of PLD in Section \ref{Sect:Improvements}). If the language contains the predicate $<$ and functions $Price, Demand$, a variant of PLD can be applied to learn laws of the form $Price(x) < Price(y) \rightarrow Demand(x) > Demand(y)$ (where $x,y$ are object variables and $Price(x) < Price(y)$ is a more convenient writing for the atom $<(Price(x), Price(y))$). Then based on the laws $Price(item_3) < Price(item_1) \rightarrow Demand(item_3) > Demand(item_1)$ and $Price(item_1) < Price(item_2) \rightarrow Demand(item_1) > Demand(item_2)$ one obtains the interval $(Demand(item_2), Demand(item_3))$ for $Demand(item_1)$, with concrete values given by $Demand$ function. 

\subsection{Anomaly Detection}
Predicates are defined to correspond to the features of interest in the input data. Those probabilistic laws learned by PLD are selected which have probability above a certain threshold (for example, these can be laws with a probability greater than 0.9). They are considered as the rules describing the normal behavior of the system and thus, violation of these rules should indicate an abnormal event. Then the task is reduced to learning a parameter corresponding to the proportion of violated rules that should indicate an anomaly.

\subsection{Time Series Analysis}
The predicate language is defined to describe past states of the given time series in terms of important features such as extreme points, technical indicators, etc. The time series data is converted into a tabular form, where each row represents predicate values for each slice of interest of the time series. Then PLD is applied to learn probabilistic laws from the resulting tabular data and the laws are used for prediction in the same way as in classification.

\subsection{Adaptive Control}
One of the PLD based approaches is close to classical Reinforcement Learning. The agent history is analyzed to infer the laws that best predict the value of the reward based on the observed state and performed (series of) action(s). The agent decides on the next action based on the rules which are applicable to the observed state (given by a set of predicates) and provide the maximal reward prediction. If no rule is applicable to the observed state, the agent makes a random action.

In the second approach, PLD is used for learning laws that predict transitions between states. The premise of every such law contains a state description in term of agent's sensor predicates and an action predicate, which corresponds to one of the available actions. The conclusion consists of predicates, which describe the state obtained after making this action. Action strategies (\emph{policies}) are learned from the transition rules by grouping them into chains. The agent is able to reason about plausibility of these policies even though some of them may represent unseen trajectories. This approach is more sample efficient than the first one, but it consists of several learning components which in general require more computing resources.

An important aspect in both approaches is that PLD allows for building complex hierarchical control systems with support for automatic subgoal discovery. An agent is able to dynamically extend the predicate language with shortcuts (new sensor predicates) for important intermediate subgoal states that must be achieved on the way to the primary goal. If a subgoal predicate is present in the premise of a transition rule used by the agent, it switches to achieving this subgoal by using the appropriate policies for the subgoal.

\section{Advantages and Limitations}
\subsection{Limitations}
\begin{itemize}
\item Probabilistic Rule Discovery is computationally expensive. The worst-case complexity of the base enumeration phase of PLD algorithm can be estimated as $N^d$, where $N$ is the number of predicates in the language and $d$ is the base enumeration depth. It is problematic to use PLD on datasets with large numbers of features (one can reduce the complexity by lowering the parameter $d$, but then the algorithm may miss many informative rules).

\item It is problematic to use PLD for datasets containing a large number of continuous features. This is due to the fact that PLD is a discrete learning model and thus, every continuous feature has to be quantized into a set of discrete ones (for example, by using one-hot coding and/or iterative splitting by medians). This, in turn, leads to problems related to the large number of discrete features and besides, with coarse splitting, some important information may be lost. In particular, because of these reasons the application of PLD for audio/video/image data is limited.

\item Similarly, it is hard to use PLD for regression tasks in general. It is required to either quantize data or predict the relationships of target feature values to the values of other features in terms of comparison predicates like $\leqslant$. 

\item In order to apply PLD, one has to explicitly define the language of predicates (features), which  essentially requires feature engineering. There exists a PLD based approach that supports feature discovery by using the so-called Probabilistic Formal Concepts \cite{b3,b4}. These combine base features into groups of interrelated ones, but anyway the set of base features must be explicitly given.
\end{itemize}

\subsection{Advantages}
\begin{itemize}
\item Probabilistic laws learned by PLD are explicit and human readable. PLD based solutions to ML tasks listed in Section \ref{Sect:Applications} are inherently interpretable and thus, PLD can be used as a basis for building explainable ML models.

\item The form of probabilistic laws resembles rules of decision trees. However, in general, for every target (predictor) predicate, the PLD algorithm learns not just a single probabilistic law, but several ones. In this aspect, PLD is closer to ensemble methods on decision trees. However, many such methods, such as, e.g., Random Forest, rely heavily on randomness in the construction of the trees and also they employ non-trivial operators for combining predictions over an ensemble. This makes these methods difficult to interpret and requires additional mechanisms for explaining outputs. In contrast, PLD has both advantages: interpretability and the feature of making decisions based on an ``ensemble'' of probabilistic laws.

\item In contrast to black-box models, the explicitness of the rule language used by PLD allows for embedding domain knowledge into the model. For example, the knowledge can be introduced as rules with probability $1$ (ground truth), or with a lower probability. When injected into the model, these rules are used for prediction together with learned probabilistic laws. 

\item The explicitness of probabilistic laws makes it possible to perform post-hoc analysis of a trained PLD model. In fact, the learned laws can be viewed as explicit instructions (an algorithm) for decision making. These can be reused for building PLD models for other datasets (transfer learning). The explicitness of the learned algorithm makes it available for meta-analysis in order to identify, for example, deeply correlated features (as implemented in the method of Probabilistic Formal Concepts), or those features/specific feature values that an object must have in order to be assigned to a particular class (abductive classification), or detection of symmetries in the structure of laws and features, which is important for sample efficiency of self-learning systems with many degrees of freedom.
\end{itemize}

\section{Directions for Improvement}\label{Sect:Improvements}
\subsection{Support for n-ary Predicates}
The choice of predicates has a direct impact on the expressiveness of PLD models. In general, PLD allows for using use not only unary, but also n-ary predicates to support laws of the form
\begin{equation}
P_1\left(x_1\right),\ldots ,P_n\left(x_n\right)\rightarrow R\left(y\right)
\end{equation}
where $x_1,\ldots ,x_n$ are (possibly non-disjoint) sets of variables and $y$ is a subset of variables from the union of $x_i$, for $i=1,\ldots , n$. Besides object variables, functional terms can be used.

Rules with n-ary predicates and functional terms are more expressive, for example, they can express relationships of features for pairs of objects like:
\begin{equation}\label{Eq:NaryRule}
Price(x) < Price(y) \rightarrow Demand(y) < Demand(x)
\end{equation}

Support for the extended rule language in PLD is currently implemented by grounding. For every combination of values for the variables of a n-ary predicate, a unary base predicate is introduced into the language. Clearly, this approach is computationally inefficient. More efficient ways to support n-ary predicates would significantly increase the expressiveness of PLD. For example, rule \ref{Eq:NaryRule} with the binary comparison predicate is an example of a regularity that is hard to express with models based on neural networks.

\subsection{Learned Quantization}
As PLD does not support continuous features, they must be converted into Boolean ones, for example, by  quantization. This implies additional difficulties as the quantization granularity needs to be carefully adjusted for each continuous feature so as to not discard information and avoid a blow-up in the number of features. Quantization can be built into the learning algorithm of PLD as follows. When enumerating rules by refinement, we first take those predicates for addition into the premise, which divide the range of the corresponding feature values by the median into two subranges. Then, after some of these predicates corresponding to a value range $r$ is added to the premise, for further refinement, we consider those predicates that split $r$ into smaller subranges, and we proceed similarly in further refinements. As a result, the range of feature values will be split up as long as the rule probability increases and its confidence is maintained. Thus, there will be no need to quantize the features in advance, the required quantization granularity will be adjusted automatically in the process of learning.

\subsection{Enumeration Reduction}
To improve the efficiency of learning, different hyperparameters are used in PLD to control the form and significance thresholds for rules (Section \ref{Sect:Hyperparameters}). The disadvantage of this approach is that it is hard to know in advance how much computational resources will be spent to search for probabilistic laws under the specified parameters. One has to make several test runs of the learning procedure to assess the complexity of computations, which is inefficient. An alternative way would be to select some limited number of the most promising rules at each level of enumeration according to some criterion. The maximal number of rules considered in the enumeration can be limited depending on the available computational resources and the selection criterion can be based, e.g, on the probability gain for rules after refinement, the entropy criterion (like in decision trees), etc.

\section{Conclusion}
In this paper, we have presented the main concepts of Probabilistic Law Discovery and discussed its advantages and limitations. Based on our analysis, we can summarize recommendations on application of PLD as follows:
\begin{itemize}
\item PLD can be used to build interpretable ML models for tasks, which require explainable  decision making
\item it is not efficient to use PLD for tasks which do not require explainability and which involve datasets with large numbers of features or which contain many continuous features (for example, image/audio/video datasets)
\item PLD can be considered as an alternative to common ML methods in tasks where explainability is not required, but where datasets contains mostly Boolean or categorical features
\end{itemize}

The principle problem of rule learning can be attacked by improving the optimization techniques already present in PLD, as well as by adopting other techniques from Operations Research and Machine Learning.  Applications of PLD to the variety of ML problems (including classification/clusterization/regression, time series/anomaly analysis, and adaptive control) evidence the high potential of models based on rule learning. We believe that elements of PLD based models can be reused to build interpetable solutions to various ML tasks on top of logic based and rule learning models like Decision Trees/Random Forests, and others.

\end{document}